# Ensemble based approach to quantifying uncertainty of LLM based classifications

**Srijith Rajamohan**
srijith.rajamohan@gmail.com
Sage

**Ahmed Salhin**
ahmed.salhin@sage.com
Sage

**Josh Frazier**
josh.frazier@sage.com
Sage

**Rohit Kumar**
rohit.kumar2@sage.com
Sage

**Yu-Cheng Tsai**
yucheng.tsai@sage.com
Sage

**Todd Cook**
todd.cook@sage.com
Sage

## Abstract

The output of Large Language Models (LLMs) are a function of the internal model's parameters and the input provided into the context window. The hypothesis presented here is that under a greedy sampling strategy the variance in the LLM's output is a function of the conceptual certainty embedded in the model's parametric knowledge, as well as the lexical variance in the input. Finetuning the model results in reducing the sensitivity of the model output to the lexical input variations. This is then applied to a classification problem and a probabilistic method is proposed for estimating the certainties of the predicted classes.

## 1 Introduction

LLMs tend to hallucinate [1] as a result of a lack of knowledge, and in this work we propose a method to measure the certainty of an LLM-generated output for a classification problem. This is crucial for key areas where trustworthy decision-making is required such as in healthcare, finance, accounting or law. There have been various efforts to solve this particular problem.

There are white-box methods and black-box methods [2] for assessing the truthfulness of LLMs. White-box techniques allow for inspection of the generation process whereas in black-box methods the model's internal states are unavailable during response generation. White-box methods rely on measuring the entropy of the generated tokens such as semantic entropy [3]. Token probability-based metrics are notable for their computational efficiency since they only a single generation is required here. The Mean Token Entropy metric described by [4] computes the average uncertainty over all the tokens in a generated response. The TokenSAR metric introduced by [5] proposes a heuristic that measures the importance of different tokens in a sequence.

Some examples of black-box methods are provided by [4], [3], [6] that incorporates techniques such as lexical similarity, number of semantic sets, sum of eigenvalues of the graph laplacian etc. Self-consistency which uses a sampling of several paths as opposed to the greedy decoding strategy [7] has been used as a way to improve the accuracy in the chain-of-thought prompting, however this approach can be sensitive to the lexical variance of the input. The Uncertainty Tripartite Testing Paradigm (Unc-TTP) metric measures LLM uncertainty as a function of the consistency of outputs as a result of the introduction of label interference [8].

The hypothesis presented in this paper posits that the variance observed in LLM outputs under a greedy sampling strategy can be attributed to the conceptual certainty encoded in the model's parameters and the variance present in the input context. Specifically, we hypothesize that the model's internal knowledge shaped through training and fine-tuning—encodes a degree of certainty, which influences how sensitive the model's outputs are to variations in the input. Fine-tuning a model can reduce the sensitivity of its output to variations in the input, resulting in more stable and reliable predictions. Finally, we present a probabilistic approach to measure the certainty of predictions for domain-specific classifications.

## 2 Methodology

LLMs can be considered to be state machines where the model state is represented by the model's parametric knowledge or the model weights. The outputs produced by a model are then a function of the parametric knowledge and the model input provided through the context. For a classification problem, we can write the following equation, where the classification can be parsed from the LLM output tokens generated until timestep $t_{\text{termination}}$:

$$\text{LLM}_{\text{classification}} = F(\text{LLM}_{\text{weights}}, \text{LLM}_{\text{input}}) \tag{1}$$

Now, we posit that the variance in the LLM-predicted class can be written as

$$\text{var}(\text{LLM}_{\text{classification}}) = F(\text{conceptual\_certainty}, \text{var}(\text{LLM}_{\text{input}})) \tag{2}$$

where the term "conceptual_certainty" represents the soundness of the parametric knowledge residing within the LLM. Ideally, fine-tuning or updating a model's weights should result in reducing the sensitivity of the predicted class in response to the lexical variance of the input, i.e. better conceptual certainty as a result of fine-tuning leads to less sensitivity to the input.

Now consider question $Q_i$ for $i \in \{0, 1, ..., k\}$, where $k$ is the space of all possible unique questions that can be asked. Consider $Q_i$ to represent the 'latent intent' which is defined as the underlying intent behind a natural language question. Now each such question can be phrased in $n$ different ways, where the same intent is expressed through a different sequence/combination of words.

$$Q_i = \{Q_{i0}, Q_{i1}, ..., Q_{in}\} \tag{3}$$

$$Q_{ij} \text{ for } i \in \{0, 1, ..., k\}, j \in \{0, 1, ..., n\} \tag{4}$$

Consider an example intent question $Q_i$ and its variants where $n = 15$.

```
Intent (Q_i) = "Who are the suppliers that I need to pay?"
```

Example variants of $Q_i$ are shown below as $Q_{ij}$.

```
Q_ij = {'Who are the suppliers that I need to pay?',
"Can you tell me the suppliers I'm currently indebted to?",
'Which suppliers have outstanding payments from me?',
'To which suppliers do I have financial obligations?',
"I'd like to know the suppliers to whom I owe money.",
'Could you list the suppliers that are awaiting payment from me?',
```

```
'Who are the vendors expecting payments from me?',
"What are the names of the suppliers I haven't paid yet?",
'Can you identify the suppliers with unpaid invoices from me?',
'To whom do I need to make payments among the suppliers?',
'Which suppliers should I settle accounts with?',
"Are there any suppliers I'm yet to pay?",
'Who are the suppliers with pending payments from my side?',
'Which suppliers am I in debt to currently?',
'Could you specify the suppliers that I need to pay?'}
```

The above can be represented as a concept space (Figure 1) where the latent intent $X$ exists (in red) and variants corresponding to this intent $X$ corresponding to $Q_i$ can be projected into this space (seen in blue). Other differing intents are shown in green. It is possible that some variants inadvertently get mapped to the incorrect intent due to a lack of knowledge within the model or ambiguity of the variant question. Through training, we can improve the number of variants (blue vectors) that maps to the correct intent $X$ (in red).

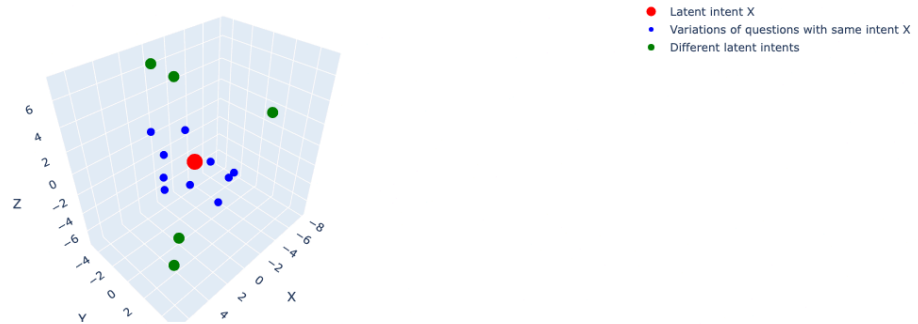

Figure 1: Concept space

In short the above can be summarized as the following:

```
Finetuning -> Improves "conceptual certainty" -> var(LLM classification) becomes less
dependent on var(LLM input)
```

## 2.1 How does it help with training?

From a concept graph perspective, it identifies the need to strengthen disambiguation either through definition or examples. Responses fall into the following three categories:

1. Confident (i.e., more certainty) and correct

2. Confident and incorrect
   - Makes a classification with high certainty but this class is wrong
   - Hard to detect
   - We should target this for improvement with training data, adding more disambiguation data

3. Not confident and incorrect
   - We can possibly detect this by building a distribution of uncertainties
   - The question could be ambiguous

- Improve through training data

Consider $Q_i$ to represent the 'latent intent' (underlying intent behind an actual natural language question).

## 2.2 What is the benefit of this approach vs. the sampling approach at inference?

It allows us to measure uncertainty of an answer relative to question intent $Q_i$ as opposed to uncertainty relative to phrasing of a question intent, i.e. $Q_{ij}$.

By setting `do_sample = True` or a variation of it to perform non-greedy sampling during the inference stage of an LLM, we can sample from the distribution of answers. With sufficient samples, we can build up the distribution of predictions and thereby compute a margin or entropy as an uncertainty measure for the predictions [9]. However, this once again computes the uncertainty for the variant $Q_{ij}$ and not the intent $Q_i$, and is therefore sensitive to the lexical formulation of $Q_{ij}$.

## 2.3 Operation of the detectors

The system under discussion is responsible for identifying the REST endpoints and the parameters necessary to form a REST route to extract the appropriate data requested by a user. Given the question `Which products are due for a reorder?`, the a fine-tuned LLM identifies the endpoints and the parameters necessary to extract the data needed to answer the question. The candidates are provided through the prompt and the LLM is asked to make a selection from the list of candidates, any results that are not present in the list are discarded. An example of the parsed dictionary from the LLM for endpoint selection is shown here:

```
`{'endpoint': '/stock_items', 'reason': 'Use the /stock_items endpoint since this
contains the purchase order reorder information <|end|>'}`
```

The response contains two keys, `endpoint` and `reason`. For the parameter detector the key would be a list of strings along with the reason key. The experiment is then conducted for the following LLMs

1. base LLM
2. trained LLM

and the distributions of predicted classes are noted and compared. If the fine-tuned LLM has learned the correct endpoint/parameters they should have a higher probability of selection relative to the base LLM.

## 3 Evaluation experiments

A few questions (179) from the space of $Q_i$ are sampled and for each question a set of n=15 variations are generated $\{Q_{i0}, ...Q_{i15}\}$ using an LLM to rephrase the original question $Q_i$. The larger the value of n the better we approximate the distribution, but it is possible to have duplicates with augmentation techniques and a larger n may not necessarily mean a larger number of unique samples. Another potential issue with this approach is the possibility of a variant deviating from the original intent when generated automatically. It must be noted that these are labeled samples.

The outputs, i.e. endpoint or parameters along with the reasoning are noted for each run and the distribution of the categorical answers (endpoints or parameters) are computed. The distribution is plotted as a bar chart with the correct answer in blue and everything else in red. The distribution of the length of the reasoning text is also plotted. We use a simple voting strategy to determine the predicted answer, i.e. the answer with the highest frequency is the predicted answer. A few terms are defined below in Table 1.

| Term | Definition |
| --- | --- |
| Ensemble predictions | $\{C0, C1, ...C14\}$ where C is a prediction from a list of endpoints or list of parameters |
| Prediction | the class that has the highest frequency |
| Ensemble_accuracy | the highest frequency (of prediction) |
| Ensemble_true_label_accuracy (blue bar) | frequency of the correct class, can be the same as ensemble_accuracy if the predicted class = correct class |

Table 1: Definitions

Consider the following example:

```
Ensemble predictions = {A, A, A, A, A, A, A, A, B, B, B, B, C, C, C}

Votes per class
A = 8, B = 4, C = 3

Prediction = A (with 8 votes out of 15)

Ensemble_accuracy = 8/15 * 100 = 53.3%

Ensemble_true_label_accuracy =  8/15 * 100 = 53.3% If the true label is 'A'
                             =  4/15 * 100 = 26.6% If the true label is 'B'
                             =  3/15 * 100 = 20%   If the true label is 'C'
```

## 3.1 Evaluating the detectors

The charts below represent the distribution of parameter or endpoint predictions for the trained model and the base model. Note that the 'accuracy' listed in the chart titles below refers to the 'ensemble_true_label_accuracy' defined above.

### 3.1.1 Evaluating the endpoint detector

Inference is performed using the base and fine-tuned models utilizing the same set of prompts. Each model is posed every variant j from the set (n=15) and the response is noted. The frequency of the predicted endpoints are plotted as a distribution for the trained model and the base model. The true label is depicted in blue while everything else is in red.

| Base model | Trained model |
|---|---|
| 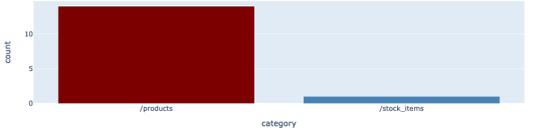 Can you list the products I hold? - /stock_items, accuracy = 6.67 | 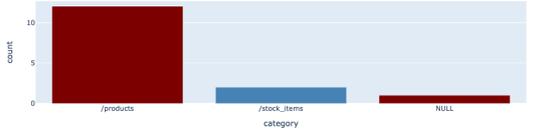 Can you list the products I hold? - /stock_items, accuracy = 13.33 |
| 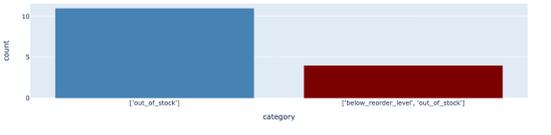 Can you tell me the items that are out of stock? - ['out_of_stock'], accuracy = 73.33 | 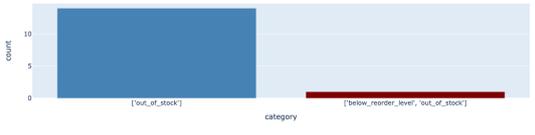 Can you tell me the items that are out of stock? - ['out_of_stock'], accuracy = 93.33 |
| 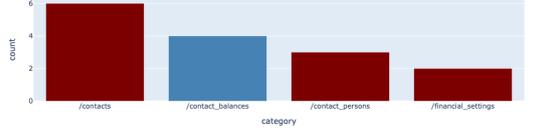 Who are my creditors? - /contact_balances, accuracy = 26.67 | 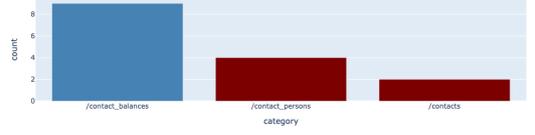 Who are my creditors? - /contact_balances, accuracy = 60.00 |
| 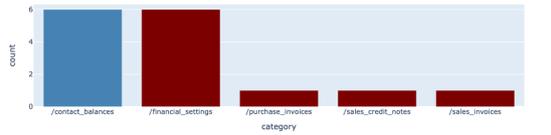 How much do I owe to my vendors? - /contact_balances, accuracy = 40.00 | 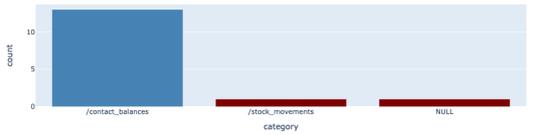 How much do I owe to my vendors? - /contact_balances, accuracy = 86.67 |
| 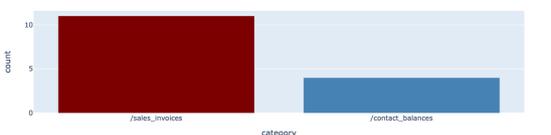 Show me my creditors with unpaid invoices - /contact_balances, accuracy = 26.67 | 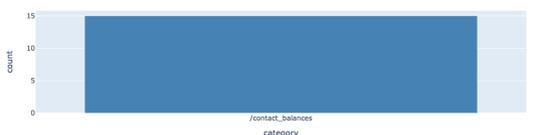 Show me my creditors with unpaid invoices - /contact_balances, accuracy = 100.00 |
| 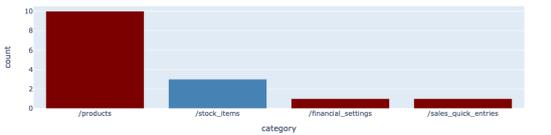 What products are available for purchase immediately? - /stock_items, accuracy = 20.00 | 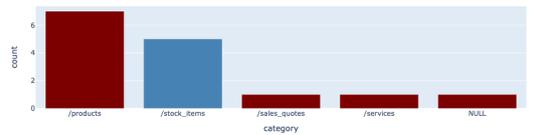 What products are available for purchase immediately? - /stock_items, accuracy = 33.33 |
| 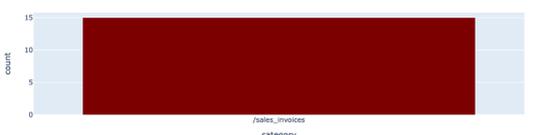 What is the sum of unpaid customer invoices? - /contact_balances, accuracy = 0.00 | 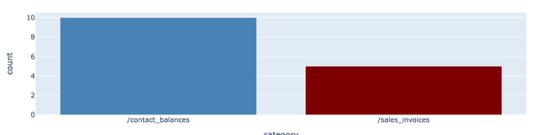 What is the sum of unpaid customer invoices? - /contact_balances, accuracy = 66.67 |

Table 2: Endpoint detector

| Base model | Trained model |
|---|---|
| 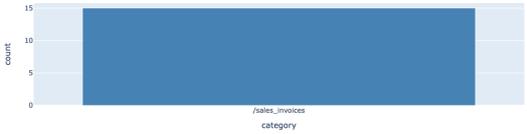 | 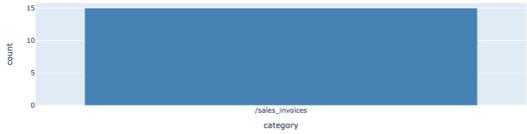 |
| 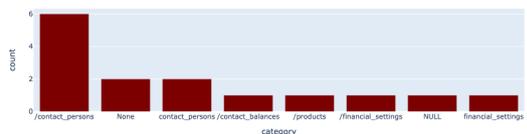 | 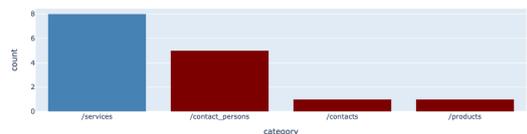 |

Table 3: Endpoint detector contd.

Note above that we generally notice an improvement in the 'ensemble_true_label_accuracy'. For example the question 'What endpoint is used to identify rates for someone?' sees an increase from 0 to 53.33% where the base model fails to recognize '/services' a viable option. Some concepts are harder to grasp, i.e. the boundaries between concepts are more nebulous and therefore poses a greater challenge for the model. It could also be the case that certain questions are more ambiguous: 'What product do I hold?' could refer to items that are being sold or items that are available for purchase immediately.

### 3.1.2 Evaluating the parameter detector

Inference is performed using the base and fine-tuned models utilizing the same set of prompts. Here the correct endpoint is known apriori and the candidate pool of parameters for the endpoint are presented to the LLMs for selection. However, unlike endpoint detection multiple parameters can be predicted for a given variant question, thereby making this a more challenging exercise since sufficiency is a stricter condition to satisfy compared to 'most likely'. Each model is posed every variant j from the set (n=15) and the response is noted.

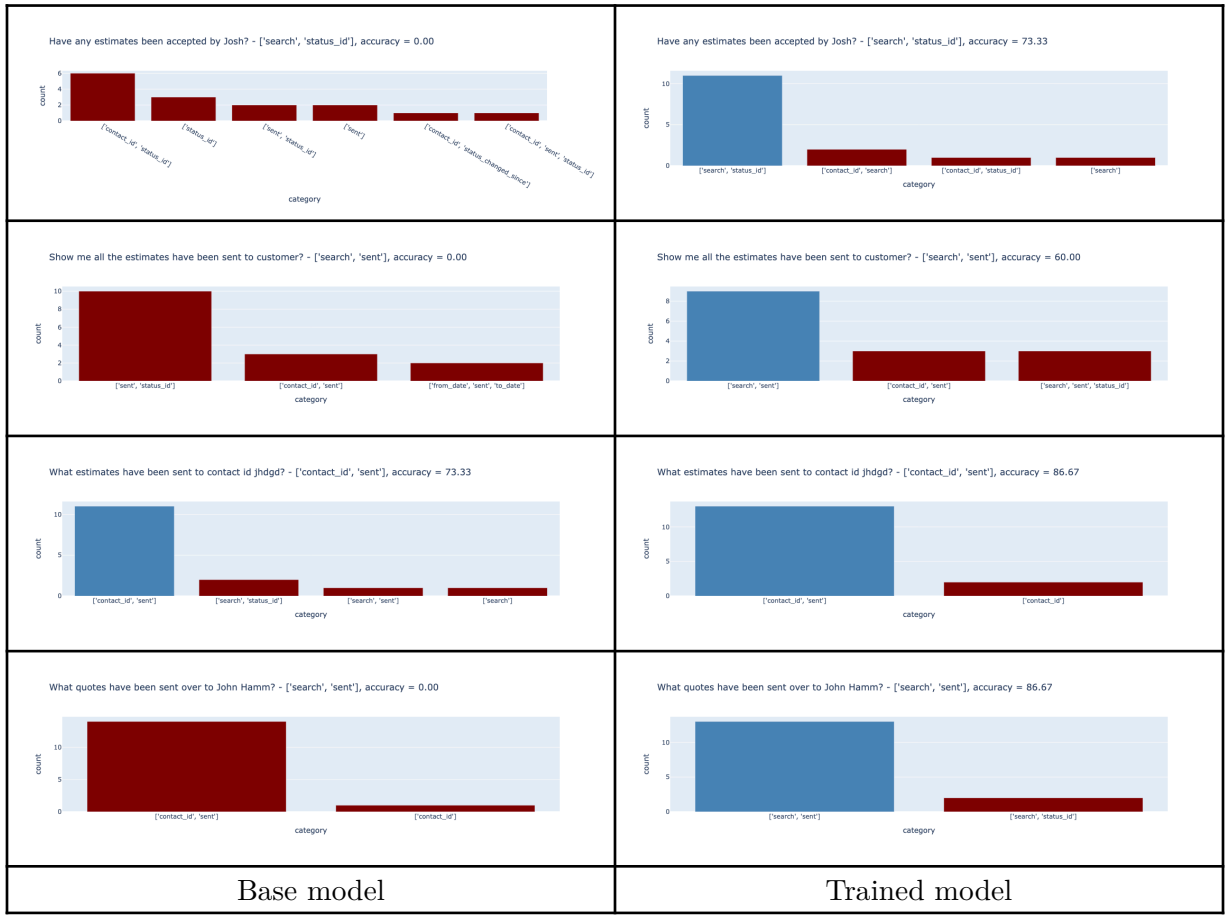

Table 4: Parameter detector

The above experiments give us some reason to believe that

1. Question phrasing variance can be used to measure uncertainty
2. Training decreases uncertainty by teaching the LLM the 'right things'

But the question remains: at inference time how do we quantify this uncertainty?

# 4 How do we quantify uncertainty?

An experiment was conducted over 179 data questions for endpoint detection. These questions are labeled and the classes span the same set of classes as in the experiments above. The prediction accuracy of endpoint detection was 92.7%. The detectors are defined such that prediction is the class which has the highest frequency from the ensemble of predictions and therefore the prediction certainty can be measured from this relative frequency or ensemble accuracy as defined above. See ensemble accuracy as shown in [Ensemble_accuracy = the highest frequency (of prediction from above)]

These certainties are curated for each of the 179 questions in our experiment and it is noted whether the prediction is correct or not based on the ground truth labels. This is divided into two sets of data `ensemble_certainty_correct` when the prediction is correct and `ensemble_certainty_incorrect` when the prediction is incorrect. This gives us two distributions for certainties (correct and incorrect predictions). The distributions of these certainties are plotted in Figure 2 below.

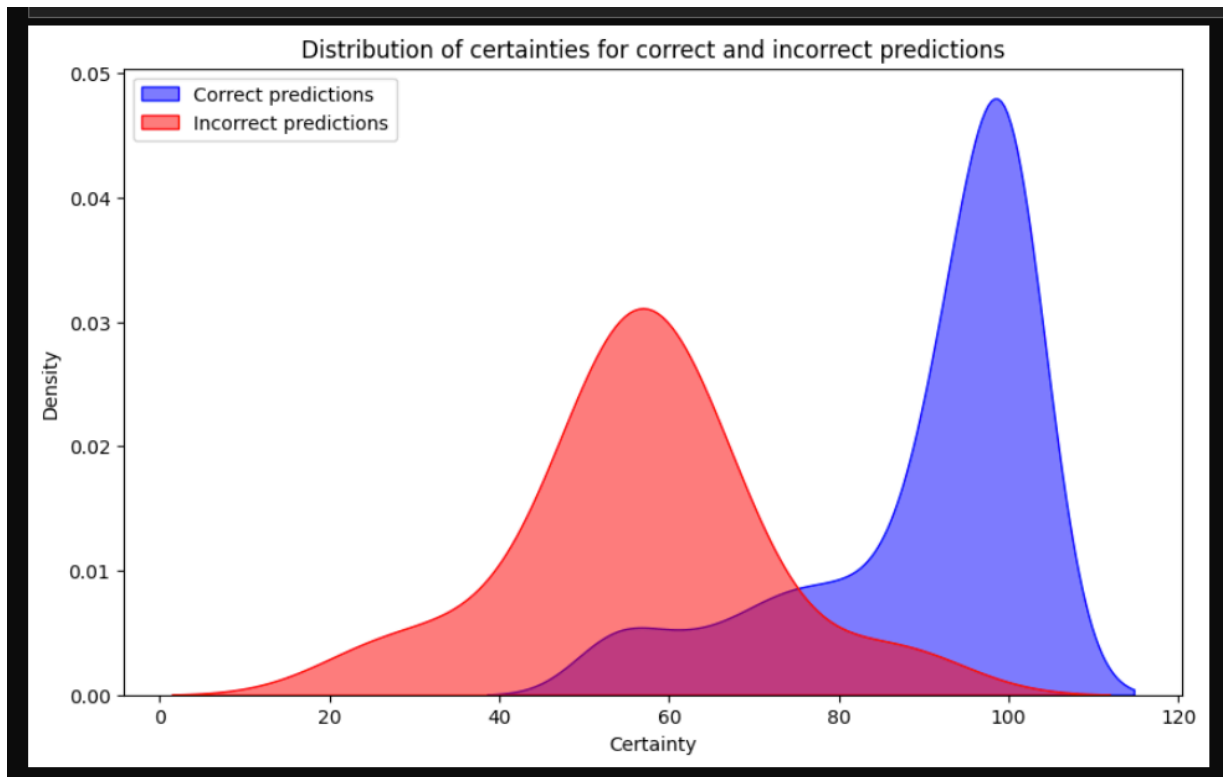

Figure 2: Distribution of certainties for correct and incorrect predictions

Note that the number of data points for `ensemble_certainty_incorrect` is relatively small and therefore a source of sampling size error. Given this data, we can determine these distributions are sufficiently different using the KS test.

```
from scipy.stats import ks_2samp

statistic, p_value = ks_2samp(ensemble_certainty_correct, ensemble_certainty_incorrect)

statistic = 0.809
p_value = 7.143e-09
```

Based on the above results there is reason to suspect that these two distributions are indeed different. Now given these prior distributions, how can we utilize it to determine whether the predicted answer to a new question is correct?

1. Perform the ensemble prediction as done above
2. Calculate the prediction and the prediction certainty using the ensemble accuracy `U_new` as above
3. Using the CDF of the 2 distributions compute the following two measures for `U_new`. These can be interpreted as measure of 'extremeness'.
   - What is the probability of seeing a value `U_new` this low from the distribution of correct values? This is computed as `P_correct(U < U_new)` `ensemble_certainty_correct_ecdf.cdf.evaluate(U_new)`
   - What is the probability of seeing a value `U_new` this high from the distribution of incorrect values? This is computed as `P(U > U_new)` `ensemble_certainty_incorrect_ecdf.sf.evaluate(U_new)`
4. Use the values to provide a measure of 'correct' or 'not correct' for the predicted class

Some examples scenarios are shown in Table 5.

| Ensemble accuracy | Probabilities | Conclusion |
|:---:|:---|:---:|
| 70% | 11.446% of correctly predicted answers have values smaller than 70<br><br>7.692% of incorrectly predicted answers have values greater than 70 | Slightly more likely to have come from the correct distribution |
| 55% | 4.819% of correctly predicted answers have values smaller than 55<br><br>46.154% of incorrectly predicted answers have values greater than 55 | More likely to have come from the incorrect distribution |
| 85% | 23.494% of correctly predicted answers have values smaller than 85<br><br>7.692% of incorrectly predicted answers have values greater than 85 | More likely to have come from the correct distribution |

Table 5: Illustrative examples

It is important to note here that this is a learning process and feedback is used to update the distributions which in turn can be used to provide more accurate estimates of correctness of the predictions.

## 5 Conclusion and Future work

There is some evidence to suggest that fine-tuning LLMs results induces greated conceptual certainty in the parametric knowledge of the LLMs making them more immune to the lexical variations of the inputs. Examples were provided to test this hypothesis using the classification problem presented here. We use an ensemble approach to estimate an 'ensemble_accuracy' to measure the prediction uncertainty and demonstrate improvement. Distributions of 'ensemble_accuracy' scores for correct and incorrect predictions are created and these prior distributions are used to compute measures that indicate the likelihood of correctness of a new prediction. While the need for more representative samples is acknowledged here, a notable area of improvement is the need for class-specific distributions. In this work distributions were built over the space of all classes but class-specific distributions are more likely to be accurate, however this requires extensive class-representative sample curation and labeling and is left for future exploration.

## Bibliography

[1] J.-Y. Yao, K.-P. Ning, Z.-H. Liu, M.-N. Ning, Y.-Y. Liu, and L. Yuan, "Llm lies: Hallucinations are not bugs, but features as adversarial examples.," *arXiv preprint arXiv:2310.01469*, 2023.

[2] E. Fadeeva *et al.*, "LM-Polygraph: Uncertainty Estimation for Language Models," 2023.

[3] L. Kuhn, Y. Gal, and S. Farquhar, "Semantic uncertainty: Linguistic invariances for uncertainty estimation in natural language generation," *ICLR*, 2023.

[4] M. Fomicheva *et al.*, "Unsupervised quality estimation for neural machine translation. ," *Transactions of the Association for Computational Linguistics, 8:539–555*, 2020.

[5] J. Duan *et al.*, "Shifting attention to relevance: Towards the predictive uncertainty quantification of free-form large language models.," *Proceedings of the 62nd Annual Meeting of the Association for Computational Linguistics*, 2024.

[6] Z. Lin, S. Trivedi, and J. Sun, "Generating with confidence: Uncertainty quantification for black-box large language models.," *CoRR*, 2023.

[7] X. Wang *et al.*, "Self-Consistency Improves Chain of Thought Reasoning in Language Models ," *arXiv:2203.11171*.

[8] H.-Y. Huang, Z. Wu, Y. Yang, J. Zhang, and Y. Wu, "Unc-TTP: A Method for Classifying LLM Uncertainty to Improve In-Context Example Selection."

[9] S. Farquhar, J. Kossen, L. Kuhn, and Y. Gal, "Detecting hallucinations in large language models using semantic entropy," *Nature*, 2024.